\documentclass[conference]{IEEEtran}
\usepackage{amsmath,graphicx,multirow,cite,setspace,array,graphics,balance}
\usepackage{hhline}
\usepackage{caption}
\usepackage{subcaption}
\usepackage[export]{adjustbox}
\usepackage{textcomp}
\usepackage{kantlipsum}
\usepackage{siunitx}
\usepackage[absolute]{textpos}
\ifCLASSINFOpdf
\else
\fi
\begin{document}
%

\begin{textblock*}{150mm}(5mm,5mm)
\noindent\normalsize A slightly different version of this paper is accepted at MMSP 2018
\end{textblock*}
\title{CPNet: A Context Preserver Convolutional Neural Network for Detecting Shadows in Single RGB Images}

\author{\IEEEauthorblockN{Sorour Mohajerani, Parvaneh Saeedi}
\IEEEauthorblockA{School of Engineering Science\\Simon Fraser University, Burnaby, BC, Canada\\
Email: \{smohajer,psaeedi\}@sfu.ca}
}


%


\maketitle

\begin{abstract}
Automatic detection of shadow regions in an image is a difficult task due to the lack of prior information about the illumination source and the dynamic of the scene objects. To address this problem, in this paper, a deep-learning based segmentation method is proposed that identifies shadow regions at the pixel-level in a single RGB image. We exploit a novel Convolutional Neural Network (CNN) architecture to identify and extract shadow features in an end-to-end manner. This network preserves learned contexts during the training and observes the entire image to detect global and local shadow patterns simultaneously. The proposed method is evaluated on two publicly available datasets of SBU and UCF. We have improved the state-of-the-art Balanced Error Rate (BER) on these datasets by 22\% and 14\%, respectively. 
    
\end{abstract}


%
\IEEEpeerreviewmaketitle

\section{Introduction}
The presence of shadows in an image is a pervasive phenomenon in scenes that are lightened by illumination sources. Shadows, on one hand, hold valuable information about scene dynamics and objects in it (e.g. detecting the buildings and vegetation regions \cite{building_vegetation} or confirming clouds locations in satellite images through their projected shadows \cite{cloud}). On the other hand, they could be a source of error and uncertainty. For instance, for tracking moving targets, those shadows projected on the background could be, erroneously, labeled as a target \cite{ moving_obj1,moving_obj2}. Therefore, identification of shadow regions in an image can considerably benefit many image processing applications including those mentioned above.

Automatic identification of shadow pixels in an image is a challenging task. The most significant complication of this problem is in how to cope with intricate mixture of the scene elements. For instance, the shape, pattern, and brightness of the shadow regions depend on many parameters including: intensity, direction, and color of the light source as well as the geometry, shape, and albedo of the obstacles. This complexity in single RGB images---unlike video frames---is duplicated since there are no temporal correlated information to benefit the identification process.

\begin{figure}[t]
\centering

\begin{minipage}{0.2\textwidth}
\centering
\centerline{\includegraphics[height=27mm, width=37mm]{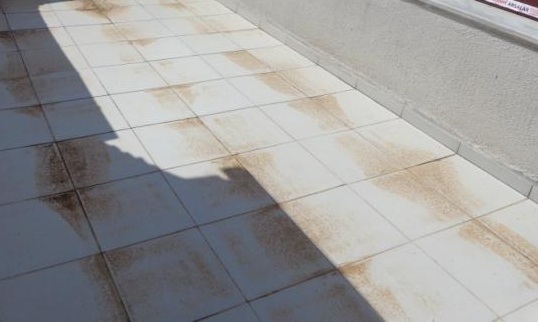}}
\end{minipage}
\vspace{1mm}
\hspace{-1mm}
\begin{minipage}{0.2\textwidth}
  \centering
  \centerline{\includegraphics[height=27mm, width=37mm]{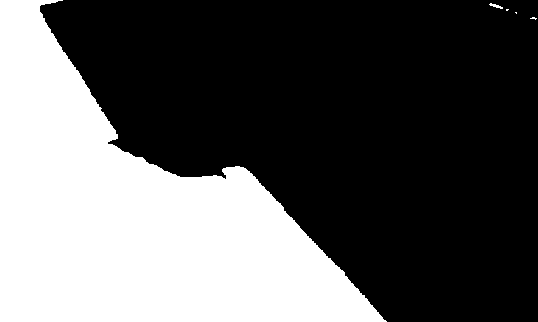}}
\end{minipage}

\begin{minipage}{0.2\textwidth}
  \centering
  \centerline{\includegraphics[height=27mm, width=37mm]{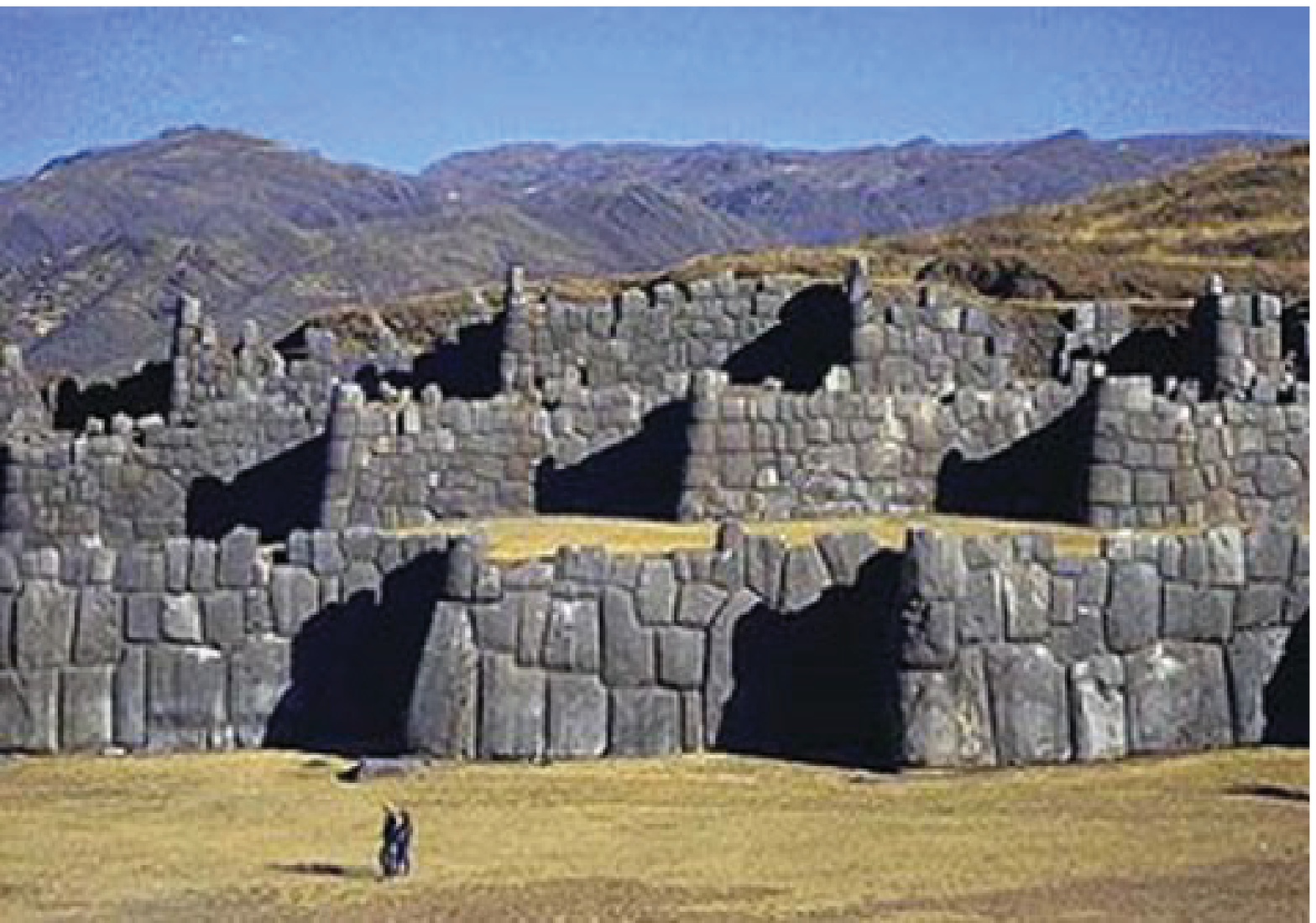}}
\footnotesize{(a)}
  \end{minipage}
\vspace{1mm}
\hspace{-1mm}
\begin{minipage}{0.2\textwidth}
  \centering
  \centerline{\includegraphics[height=27mm, width=37mm]{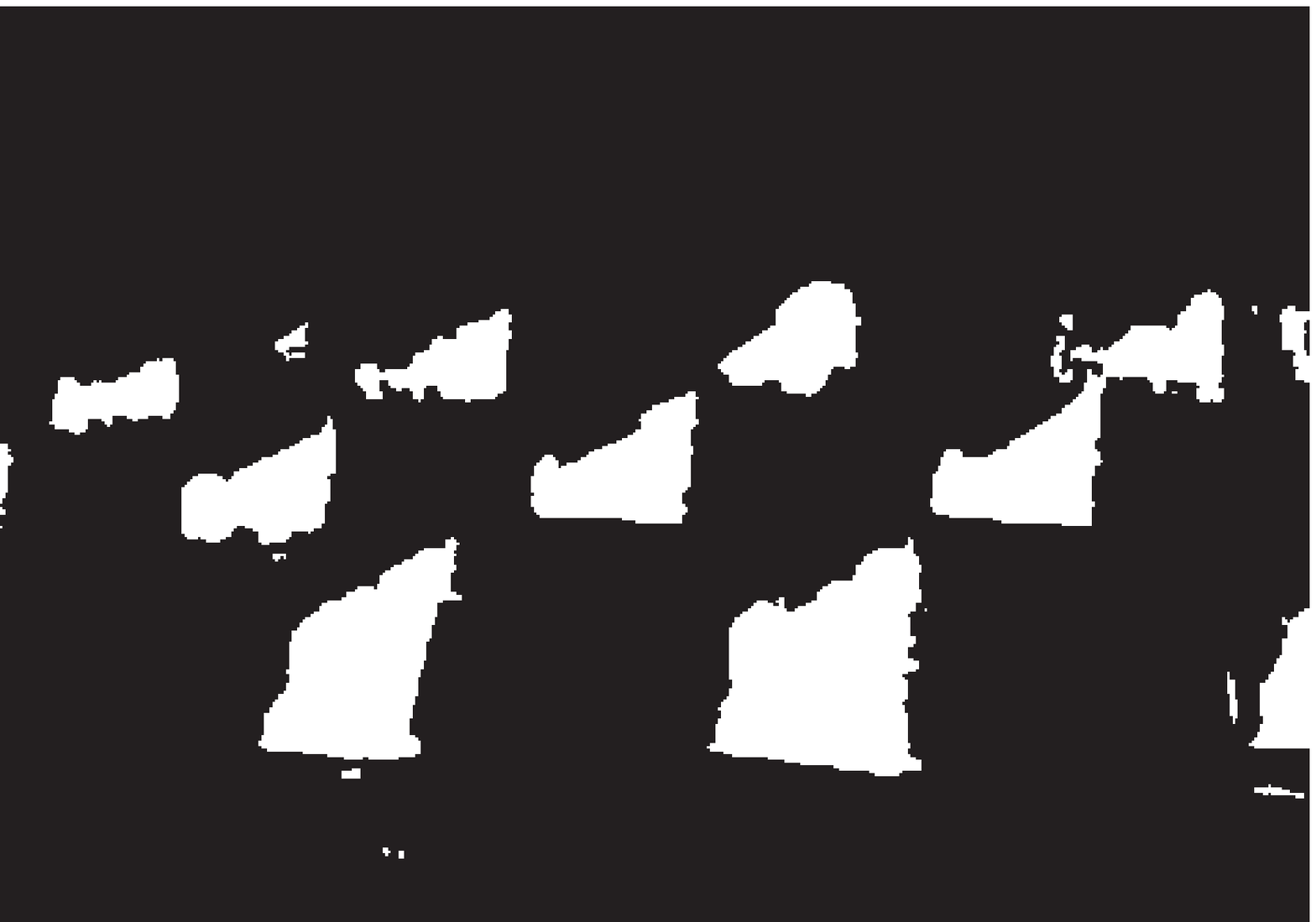}}
\footnotesize{(b)}
  \end{minipage}
\vspace{-1mm}
\hspace{-2.5mm}  
\setlength{\abovecaptionskip}{2mm}
\caption{\small Examples of shadow masks obtained by the proposed method: (a) input RBG images, (b) predicted masks\label{Fig:just_samples}}
\end{figure}

\begin{figure*}[t]

\begin{minipage}{1\textwidth}
  \centering
  \centerline{\includegraphics[ width = 160mm]{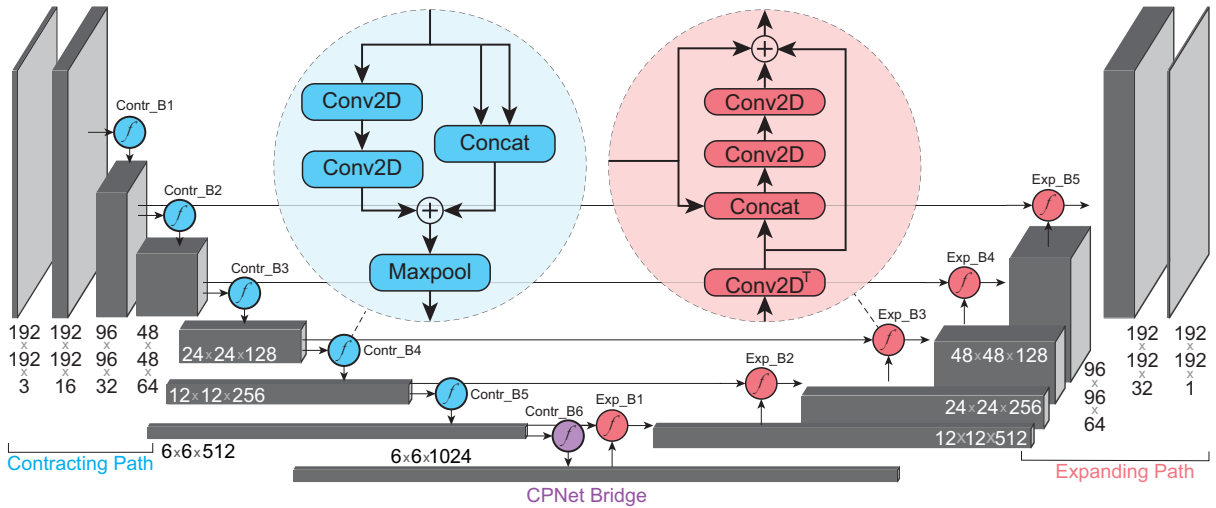}}{\vspace{2mm}}

\end{minipage}

\caption{\small CPNet architecture. The Conv2D block is a convolution layer with stride=1, the Maxpool is a max-pooling layer with pool size=2, the Concat block applies a depth-wise concatenation, and Conv2D\textsuperscript{T} is a convolution transposed layer with stride=2.
\label{Fig:Arch}}

\end{figure*}

In recent years, many researches have been developed to address the shadow detection problem in a single image. These methods are, mostly, divided into two categories: handcrafted \cite{handcrft1,UCF,UIUC_jornl} and deep-learning based methods \cite{khan,SBU_dataset,scGAN}. Unfortunately, the former requires a complete understanding of all the above mentioned elements for a precise labeling of the shadow regions. This is not possible in most of the actual cases since there is no prior information about the light source(s) or the blocking object(s). Although proposed deep-learning based methods have been more successful than the handcrafted ones, they could suffer from the lack of robustness in extracting sufficient global and local shadow contexts. Consequently, their final results are not as good as one may expect or require.

Here, we propose a deep-learning based approach, which has a simple yet efficient structure for training and prediction. The skeleton of this network is based on U-Net\cite{UNet}. U-Net is widely acknowledged as one of the best CNN architectures for semantic segmentation purposes \cite{3DUNet,finger_prnt_UNet}. It takes advantage of low-level features in a contracting path to construct the high-level features of an expanding path. ResNet \cite{resnet} is another successful modern architecture, which is specifically designed for the classification purposes \cite{resnet_apps1}. It, basically, takes advantage of some specific connections to manage one of the optimization problems of deep CNNs---gradient vanishing---during the training phase. The effect of those specific connections, also, can be interpreted that by adding feed-forward connections between two convolution layers, the forgotten features of the first layer are taken into account in the second one. As a result, the learned semantic features that are transferred to the next layer are superior. In this work, these two characteristics from two modern CNN architectures are combined. The main contribution of the proposed network in this paper is to utilize the learned contexts of the image and preserve them for retrieval of shadow patterns, which leads to an improvement of the BER measure. Our modifications add negligible amount---around 5\%---of complexity (parameters) to the baseline model. No prior information, tuning of the parameters, or any pre/post processing are required in the proposed method. Fig. \ref{Fig:just_samples} illustrates two examples of predicted shadow masks by the proposed method.

Image segmentation approaches based on deep learning require a large number of images with their pixel-level annotated ground truth for training. The Stony Brook University (SBU) shadow dataset \cite{SBU_dataset} is the largest publicly available dataset consisting of 4085 images for training and 638 images for test purposes. It covers a wide range of scene conditions and elements such as dim and bright lights, cast and self-shadows, indoor and outdoor scenes, low and high quality pictures, etc. Another shadow database by the University of Central Florida (UCF) consists of 355 images from which 245 randomly selected images are used for shadow detection \cite{UCF, khan}. The proposed CNN is trained on the SBU training set and is evaluated on SBU and UCF test sets. It has improved the Balanced Error Rate (BER) by 22\% on SBU and 14\% on UCF test sets, respectively.

\section{Related Work}
Many methods were developed to identify shadows in a single RGB image. Traditional approaches are, mostly, built on the fact that the illumination component of shadow pixels are different from non-shadow ones \cite{UCF,UIUC_jornl}. Vicente et al. \cite{leave_one_out} proposed a statistical learning-based method, which first segmented the image into multiple regions and then labeled each of the regions as shadow and non-shadow using a Least Squares Support Vector Machine (LSSVM). Since 2016, modern deep-learning based approaches have been proposed to address the problem of shadow detection. Khan et al. \cite{khan} introduced ConvNet method, which combined a Conditional Random Field (CRF) model with two CNNs to extract the local features of the shadow pixels in an image. Later, Vicente et al. \cite{SBU_dataset}, in their proposed method (Stacked-CNN), utilized the learned semantic features from a prior CNN to train another patch-wise CNN and refine the details of shadow regions in an image. It is worth mentioning that the authors of \cite{SBU_dataset}, considerably, paved the way of using deep-learning based methods in shadow detection field by introducing the largest shadow detection dataset of single RGB images---SBU dataset. Recently, Nguyen et al. \cite {scGAN} proposed a Conditional Generative Adversarial Network (scGAN). This framework benefited from a generator and a discriminator network, which jointly worked to identify shadow masks in an image.  

In this paper, we introduce Context Preserver convolutional neural Network (CPNet), which takes advantage of feed-forward connections to isolate shadow pixels. Unlike the ConvNet and Stacked-CNN, the proposed method is not computationally expensive during the training and the prediction, since it only has one set of weights. Besides, both ConvNet and Stacked-CNN are patch-based methods and therefore they are not capable of capturing the shadow features from the entire image and, consequently, fail to observe the existing global semantic contexts in an image. Unlike the scGAN, the training procedure of the CPNet is straight-forward, therefore, it can be simply adapted for other image segmentation tasks. Additionally, it does not require adjusting of a sensitivity parameter at the prediction phase \cite{scGAN}.  

\section{Proposed Method}
The proposed network is built on the U-Net. The block diagram of the CPNet is shown in Fig. \ref{Fig:Arch}. Overall, it has eleven convolution blocks in both paths (contr\_B1, ..., contr\_B6, exp\_B1, ..., exp\_B5) and a bridge block, which connects these two paths together. In the contracting path (the left side of the architecture), the input of a convolution block is followed by two convolution layers, a summation block, and a max-pooling layer. The convolution layers encode the low-level characteristics of the input, while the pooling layer reduces the spatial size of the features. The main contribution of the summation block is to add a specific version of the input to the output to include more low-level image attributes. This specific version is a duplication of the input---concatenation of the input with itself. Moreover, the connection from the input of the block to the output, in this deep neural network, avoids gradient degrading during back propagation procedure. The most important contexts of the image, which are gathered in the bridge block, are gradually decoded to the full resolution using five blocks of the expanding path. In each of these blocks, there are one convolution transposed layer and two convolution layers following by a summation. In contrast to the recent modified versions of the U-Net \cite{resUNet,scGAN}, the CPNet utilizes the analogous features of the contracting path for two purposes: first, reconstruction of the contexts inside of each block of the expanding path and second, transferring these features to the next block. This way, the network is forced to preserve---and transfer---the learned contexts in almost every layer of the network. Therefore, the ability of the network to remember features is improved and, as a result, convergence to a better local minimum is achieved.

\subsection{Training}
The spatial size of the input of the network is $192\times192$ pixels. Before training, the entire image is resized to $192\times192$ to fit the network. The activation function of all of the convolution layers as well as summation blocks is \textit{Rectified Linear Units (ReLU)} \cite{ReLU}, which is followed by a batch normalization layer \cite{batchnorm} to speed up the learning process. The size of the kernels in all the convolution layers is $3\times3$. There is one \textit{dropout} layer \cite{dropout} in the Contr\_B6 block with dropping rate of $0.15$. The usage of dropout layer right after the deepest convolution layer, which has the largest number of parameters, prevents the network from overfitting. A convolution layer with \textit{sigmoid} activation function extracts the output pixel-wise probabilities.

The network learns the shadow features while it minimizes the following soft \textit{Jaccard} loss function \cite{jacc1, jacc2}:
\begin{equation}
\large
\begin{split}
J_{loss}(y_t,y) \! = \!-\dfrac{\sum\limits_{i=1}^{n} y_{t_i} y_i+\epsilon}{\sum\limits_{i=1}^{n} y_{t_i} + \sum\limits_{i=1}^{n} y_i - \sum\limits_{i=1}^{n} y_{t_i} y_i+\epsilon},
\\ 
\end{split}
\label{Eq:loss}
\end{equation}
where $y_t \in \{0,1\}$ denotes the provided ground truth and $y \in [0,1]$ represents the obtained probabilities by the network. Since $y_t$ and $y$ have the same size ($H \! \! \times \! \! W$), we set $n = H \! \! \times \! \! W$. $y_i$ and $y_{t_i}$ are the $i$th pixel of $y$ and $y_t$ respectively. $\epsilon$ is a small positive real number to avoid division by zero in images with no shadow pixels.

The initial learning rate for the training of the CPNet is $10^{-4}$. Adam optimizer \cite{ADAM} is used for gradient descending of the loss function. Geometric data augmentation---such as horizontal flipping, rotation (with angle range of [\ang{-20}, \ang{20}]), and zooming (with scale range of [1.2,2.5]) is utilized to improve the regularization and generalization aspects of the approach by forcing the network experiencing a small amount of randomness in each of the training epochs.  

\subsection{Prediction (Test)}
After the training, we end up with one set of weights, which is sufficient for prediction purposes. A multi-scale scheme is used to obtain the most important features out of the learned shadow patterns. We, first, resize the image to four sizes of $192\times192$, $256\times256$, $384\times384$, and $480\times480$. Then, the network is fed with these different scaled images. Therefore, four probability maps are obtained out of each input image. By resizing these outputs to the original input size and using an ensemble approach, all probability maps are merged together to create a binary mask. To ensemble the maps two methods are inspected:
\begin{itemize}
  \item Calculating the average of four probability maps, then applying a simple thresholding on the average result.
  \item Thresholding of four probability maps to create a binary mask for each of them, then applying a pixel-wise logical $O \! R$ on the four obtained binary masks.
  \end{itemize}
Since the latter showed a slightly lower BER measure, it was used during the prediction phase.
\begin{figure*}[t]
\centering

\begin{minipage}{0.15\textwidth}
\centering
\centerline{\includegraphics[height=20mm, width=30mm]{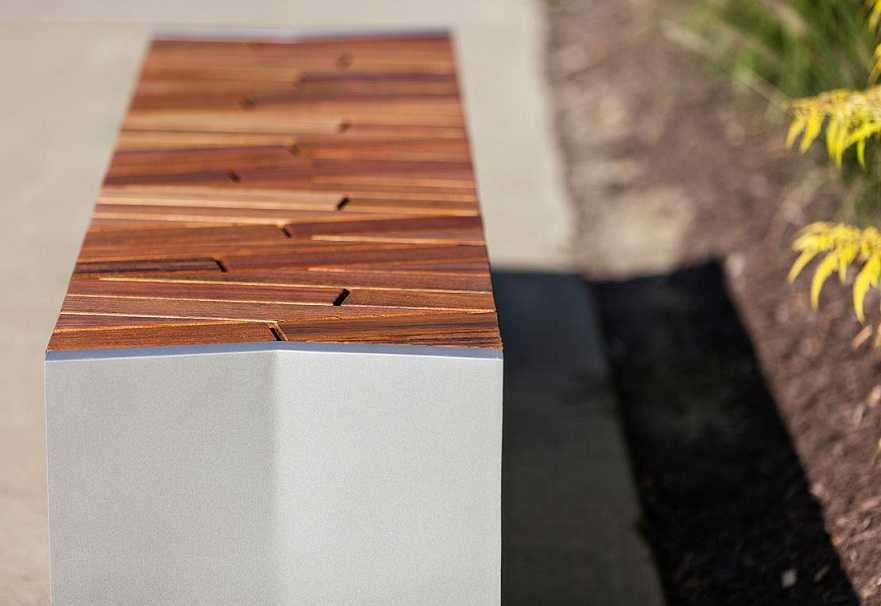}}

\end{minipage}
\vspace{1mm}
\hspace{1mm}
\begin{minipage}{0.15\textwidth}
  \centering
  \centerline{\includegraphics[height=20mm, width=30mm]{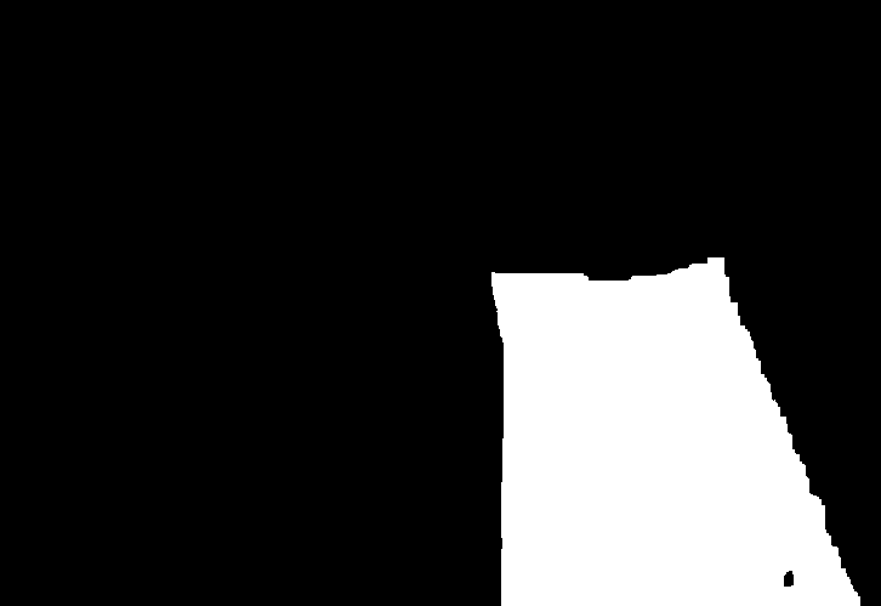}}

\end{minipage}
\vspace{1mm}
\hspace{1mm}
\begin{minipage}{0.15\textwidth}
  \centering
  \centerline{\includegraphics[height=20mm, width=30mm]{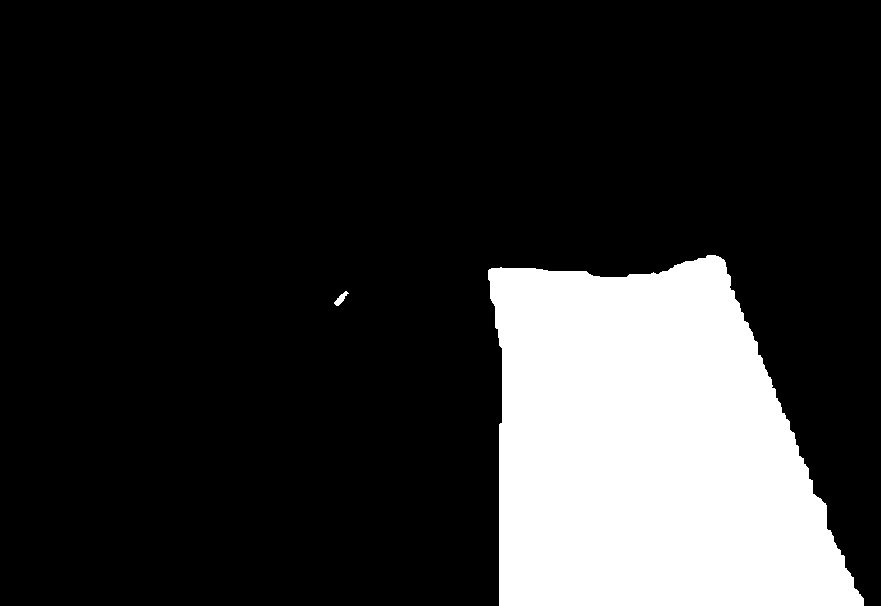}}

\end{minipage}
\vspace{1mm}
\hspace{2mm}
\begin{minipage}{0.15\textwidth}
\centering
\centerline{\includegraphics[height=20mm, width=30mm]{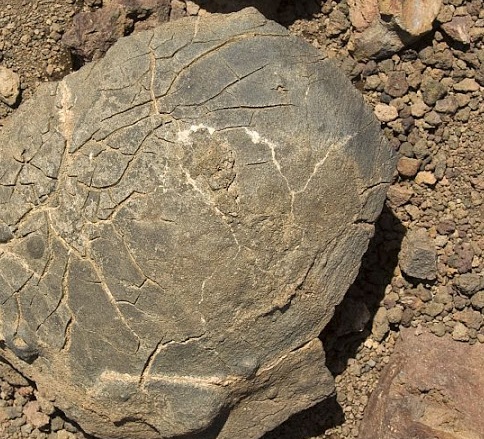}}

\end{minipage}
\vspace{1mm}
\hspace{1mm}
\begin{minipage}{0.15\textwidth}
  \centering
  \centerline{\includegraphics[height=20mm, width=30mm]{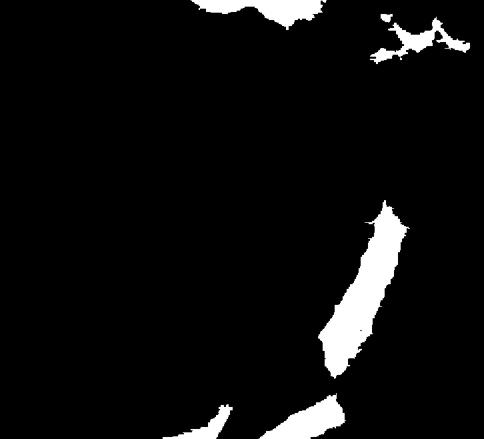}}

\end{minipage}
\vspace{1mm}
\hspace{1mm}
\begin{minipage}{0.15\textwidth}
  \centering
  \centerline{\includegraphics[height=20mm, width=30mm]{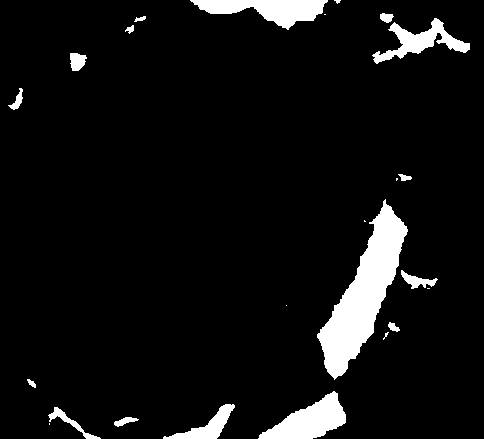}}

\end{minipage}
\begin{minipage}{0.15\textwidth}
  \centering
  \centerline{\includegraphics[height=20mm, width=30mm]{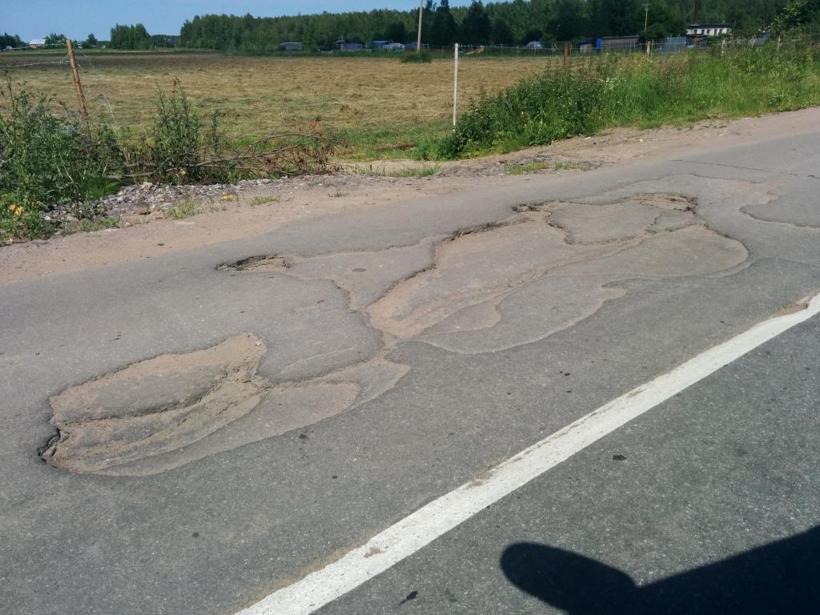}}
\footnotesize{(a)}
  \end{minipage}
\vspace{1mm}
\hspace{1mm}
\begin{minipage}{0.15\textwidth}
  \centering
  \centerline{\includegraphics[height=20mm, width=30mm]{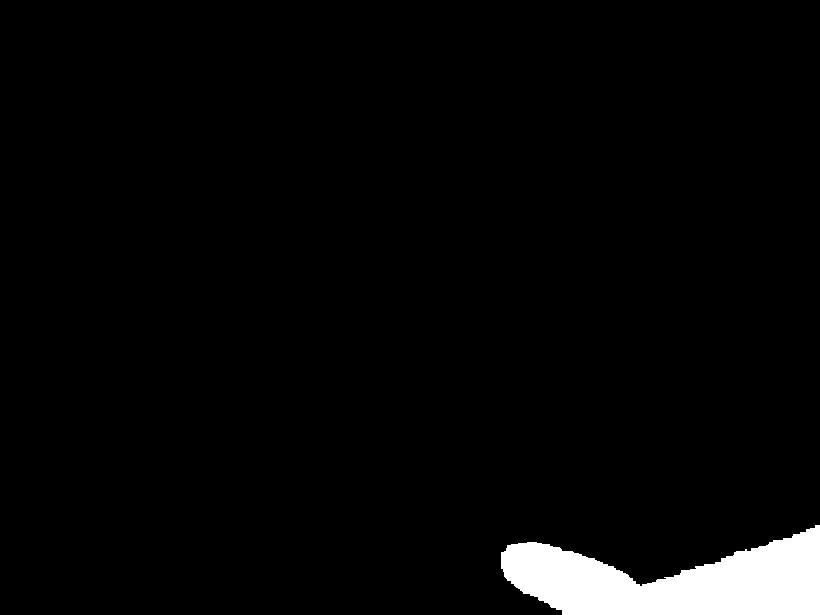}}
\footnotesize{(b)}
  \end{minipage}
\vspace{1mm}
\hspace{1mm}
\begin{minipage}{0.15\textwidth}
  \centering
  \centerline{\includegraphics[height=20mm, width=30mm]{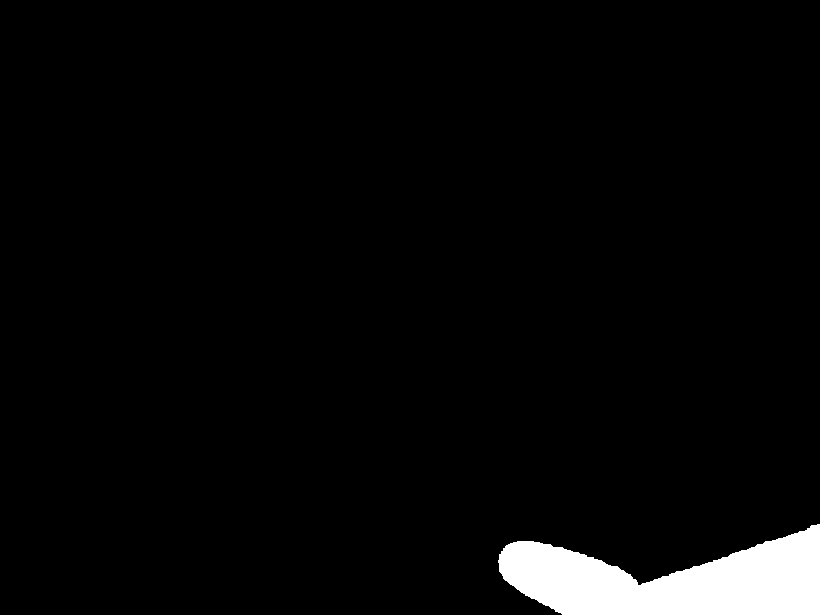}}
\footnotesize{(c)}
\end{minipage}
\vspace{1mm}
\hspace{2mm}
\begin{minipage}{0.15\textwidth}
  \centering
  \centerline{\includegraphics[height=20mm, width=30mm]{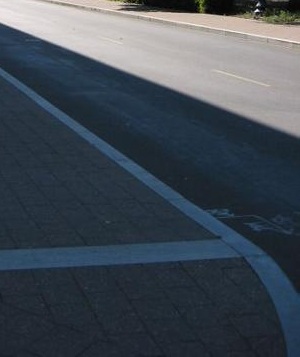}}
\footnotesize{(d)}
  \end{minipage}
\vspace{1mm}
\hspace{1mm}
\begin{minipage}{0.15\textwidth}
  \centering
  \centerline{\includegraphics[height=20mm, width=30mm]{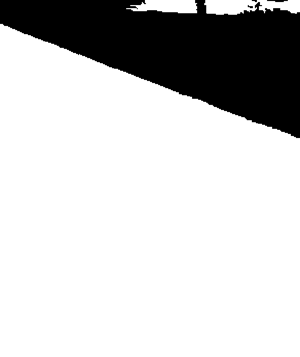}}
\footnotesize{(e)}
  \end{minipage}
\vspace{1mm}
\hspace{1mm}
\begin{minipage}{0.15\textwidth}
  \centering
  \centerline{\includegraphics[height=20mm, width=30mm]{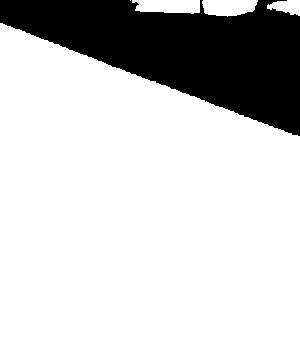}}
\footnotesize{(f)}
\end{minipage}
\vspace{-1mm}
\hspace{-2.5mm}  
\setlength{\abovecaptionskip}{2mm}
\caption{\small Examples of the shadow masks obtained by the proposed method: (a),(d) input RBG images, (b),(e) corresponding ground truths, (d),(f) predicted masks\label{Fig:experimental_samples}}
\end{figure*}

\section{Experimental Results}

\subsection{Datasets}
\subsubsection{Datasets for training}
The CPNet is trained from scratch on the SBU shadow training set that includes of 4085 images. Unlike the UCF dataset---which the corresponding ground truth of the images are manually annotated---the ground truth for SBU images are obtained using a semi-automatic approach \cite{SBU_dataset}. As this dataset does not overlap with the UCF dataset, it is worth training the same network on the UCF training set, too. Despite the fact that the exact subset of training data for this dataset is not reported in \cite{UCF}, we tried to train the proposed network on a randomly selected training set similar to \cite{UCF, SBU_dataset, scGAN}. Unfortunately, the limited number of images of training set---125 images---did not allow the CPNet to converge to an appropriate point. Therefore, we trained CPNet only on the SBU training set.
\subsubsection{Datasets for test}
The CPNet is evaluated on 638 images of the SBU test set and 120 images of the UCF test set. The exact test set of the UCF dataset used in \cite{UCF} is not known. To perform a fair comparison of our results on the UCF test set with the state-of-the-art results, we considered the following test sets:
\begin{itemize}
  \item Five subsets of 120 test images are independently and randomly selected similar to \cite{UCF}. The average quantitative results of these subsets are reported under \textit{Average UCF*}.
   \item The numeric results over all 355 images of the UCF dataset are reported under \textit{All 355 UCF}.
\end{itemize}

\subsection{Evaluation Metrics}

As image segmentation is a pixel-level mapping, it requires a validation metric to consider the unbalanced distribution of the pixels in different classes in the output images. We use the BER measure to evaluate the performance of the proposed method:
 
\begin{equation}
\normalsize
\vspace{2mm}
BER = 1 -\dfrac{1}{2} \big( \dfrac{T \! P}{T\! P+F\! N} + \dfrac{T\! N}{T\! N+F\! P} \big),
\\ 
\label{Eq:Evber}
\end{equation}
where $T \! P$, $T \! N$, $F \! P$, and $F \! N$ are the number of truly predicted shadow pixels, truly predicted non-shadow pixels, falsely predicted shadow pixels, and falsely predicted non-shadow pixels, respectively. Also, Per Pixel Error Rate (PER) of shadow and non-shadow classes are reported in our work.

\renewcommand{\arraystretch}{1.1}
\begin{table}
\small
\begin{minipage}[t]{0.5\textwidth}
\centering
\caption{Perfromance of the proposed method on the SBU test set (in~\%).\label{Tab:SBU}} 

\begin{tabular}{|c|c|c|c|}
\hline
\multirow{2}{*}{\textbf{Methods}} & \multirow{2}{*}{\textbf{BER}} & \multicolumn{2}{c|}{\textbf{PER} }       \\ 
\cline{3-3}\cline{4-4}
                        &                          & \textbf{Shadow}                       & \textbf{Non-Shadow}  \\ 
\hhline{|=|=|=|=|}
Stacked-CNN \cite{SBU_dataset}      & 11.0         & 9.6             & 12.5 \\  \hline
scGAN \cite{scGAN}           & 9.1          & \textbf{7.8}             & 10.4 \\ \hline
CPNet (This Paper)             & \textbf{ 7.1 }         & 9.0            & \textbf{5.2}                      \\ \hline
\end{tabular}
\end{minipage}
\end{table}

\renewcommand{\arraystretch}{1.1}
\begin{table}
\small
\begin{minipage}[t]{0.5\textwidth}
\centering
\caption{Performance of the proposed method on the UCF test set (in~\%).
\label{Tab:UCF}} 
\setlength\tabcolsep{4pt} 
\begin{tabular}{|>{\centering\arraybackslash}m{2cm}|>{\centering\arraybackslash}m{1.3cm}|c|c|c|}
\hline
\multirow{2}{*}{\textbf{Methods}} & \multirow{2}{*}{\textbf{Test Set}} & \multirow{2}{*}{\textbf{BER}} & \multicolumn{2}{c|}{\textbf{PER} }       \\ 
\cline{4-4}\cline{5-5}
                        &                         &  & \textbf{Shadow}                       & \textbf{Non-Shadow}  \\ 
\hhline{|=|=|=|=|=|}
Stacked-CNN \cite{SBU_dataset}       & UCF   & 13.0          & 9.0             & 17.1                \\ \hline
scGAN \cite{scGAN}             & UCF                     & 11.5          & \textbf{7.7}    & 15.3                \\ \hline
CPNet (This Paper) & Average UCF* & \textbf{9.9}           & 11.7            & 8.1         \\  \hhline{|=|=|=|=|=|}
ConvNet \cite{khan}           & All 355 UCF             & 14.7          & 22.0            & \textbf{7.4}                 \\ \hline
CPNet (This Paper) & All 355 UCF             & \textbf{10.1} & \textbf{11.7}   & 8.5        \\ \hline
\end{tabular}
\end{minipage}
\end{table}
    
\subsection{Comparison with the state-of-the-art}
Table \ref{Tab:SBU} shows experimental results on SBU test set. In terms of BER measure, our method outperforms the state-of-the-art method---scGAN \cite{scGAN}---by 22\%. This is a significant amount, especially, when one considers the fact that this test set consists of 638 images with different types of shadows and scenes. Additionally, the PER measure of the non-shadow pixels is reduced by 50\% compared to scGAN, which highlights the robustness of the proposed method in not overlabeling pixels as shadow. Moreover, the shadow pixels PER measure is reduced by 6.2\% in comparison to Stacked-CNN \cite{SBU_dataset}. Our reported improvement in the numeric results are calculated by: \textit{100$\times$(old\_measure-new\_measure)/old\_measure}. Some visual examples of the predicted shadow masks from sample images of the SBU test set are displayed in Fig. \ref{Fig:experimental_samples}.

Table \ref{Tab:UCF} summarizes the performance of our system on UCF test set. The average BER measure of five randomly selected UCF test sets is 9.9\%, which is 14\% better than the state-of-the-art's BER measure. The BER on prediction results of all of the 355 images of the UCF dataset is also calculated. The CPNet improves this BER measure 31\% compared to the ConvNet \cite{khan}. It is worth noting that the BER measure of the state-of-the-art handcrafted method \cite{leave_one_out} on the UCF test set is 13.2\%.

%
%
%
%
%

\section{Conclusion}
In this paper a context preserver CNN is introduced to address the problem of shadow detection in single RGB images. The proposed method preserves learned features of the image and retrieves the local and global contexts of the image. This approach, which does not require any pre/post processing, can be exploited in many computer vision applications. In addition, since its implementation is simple, it can be adapted in any other image segmentation algorithm. The proposed method outperforms the state-of-the-art approaches by a wide margin. Our future work on this topic is to improve the performance on images with dim light.



\balance


\bibliographystyle{IEEEtran}
\bibliography{refs}

\end{document}